\documentclass{article}

\usepackage[numbers]{natbib}
\usepackage[preprint]{neurips_2025_custom}

\usepackage{amsmath,amsfonts,bm}

\def\eqref#1{equation~(\ref{#1})}

\def\1{\bm{1}}

\DeclareMathAlphabet{\mathsfit}{\encodingdefault}{\sfdefault}{m}{sl}
\SetMathAlphabet{\mathsfit}{bold}{\encodingdefault}{\sfdefault}{bx}{n}

\usepackage[dvipsnames]{xcolor}         
\definecolor{linkColor}{rgb}{0.18,0.39,0.62}
\usepackage[utf8]{inputenc} 
\usepackage[T1]{fontenc}    
\usepackage[colorlinks=true,linkcolor=linkColor,citecolor=linkColor,filecolor=linkColor,urlcolor=linkColor]{hyperref}       
\usepackage{cleveref}
\usepackage{multirow}
\usepackage{xspace}
\usepackage{booktabs}
\usepackage{graphicx}
\usepackage{array}
\usepackage{wrapfig}
\usepackage{bm}
\usepackage{algorithm}

\usepackage{algpseudocode}
\usepackage{enumitem}
\usepackage{tcolorbox}
\usepackage{makecell}
\usepackage{diagbox}

\usepackage{amsmath}
\usepackage{amssymb}
\usepackage{mathtools}
\usepackage{amsthm}

\usepackage{multirow}
\usepackage{amsmath}
\usepackage{capt-of}
\usepackage{tabularx}
\usepackage{epsfig}
\usepackage{amssymb}
\usepackage{amsfonts}
\usepackage{booktabs}
\usepackage{scalerel}
\usepackage{listings}
\usepackage{varwidth}
\usepackage{stmaryrd}
\usepackage{bbm}
\usepackage{wrapfig}
\usepackage{pifont}

\newcommand{\tabincell}[2]{\begin{tabular}{@{}#1@{}}#2\end{tabular}}

\definecolor{deepblue}{rgb}{0,0,0.5}
\definecolor{officeblue}{RGB}{0,102,204}
\definecolor{deepred}{rgb}{0.6,0,0}
\definecolor{deepgreen}{rgb}{0,0.5,0}
\definecolor{mybrickred}{RGB}{182,50,28}

\definecolor{fillcolor}{RGB}{216,217,252}

\usepackage{etoolbox}
\usepackage{framed}

\newif\ifxetexorluatex
\ifxetex
  \xetexorluatextrue
\else
  \ifluatex
    \xetexorluatextrue
  \else
    \xetexorluatexfalse
  \fi
\fi
\newcommand*\quotesize{60} 
\newcommand*{\openquote}
   {\tikz[remember picture,overlay,xshift=-4ex,yshift=-2.5ex]
   \node (OQ) {\fontsize{\quotesize}{\quotesize}\selectfont``};\kern0pt}

\newcommand*{\closequote}[1]
  {\tikz[remember picture,overlay,xshift=4ex,yshift={#1}]
   \node (CQ) {\fontsize{\quotesize}{\quotesize}\selectfont''};}

\colorlet{shadecolor}{white}

\newcommand*\shadedauthorformat{\emph} 

\newcommand*\authoralign[1]{%
  \if#1l
    \def\authorfill{}\def\quotefill{\hfill}
  \else
    \if#1r
      \def\authorfill{\hfill}\def\quotefill{}
    \else
      \if#1c
        \gdef\authorfill{\hfill}\def\quotefill{\hfill}
      \else\typeout{Invalid option}
      \fi
    \fi
  \fi}
{\authoralign{#1}
\ifblank{#2}
   {\def\shadequoteauthor{}\def\yshift{-2ex}\def\quotefill{\hfill}}
   {\def\shadequoteauthor{\par\authorfill\shadedauthorformat{#2}}\def\yshift{2ex}}
\begin{snugshade}\begin{quote}\openquote}
{\shadequoteauthor\quotefill\closequote{\yshift}\end{quote}\end{snugshade}}

\newcommand{\github}{\raisebox{-1.5pt}{\includegraphics[height=1.05em]{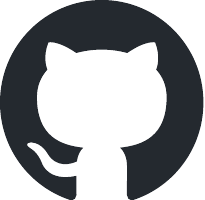}}\xspace}

\definecolor{DarkBlue}{RGB}{0, 51, 153}
\newcommand{\ours}{EL}

\title{LLM-as-a-Coach: Experiential Learning for Non-Verifiable Tasks}

\author{%
Tianzhu Ye\thanks{~Equal contribution.}~~~~~~~~Li Dong\footnotemark[1] \\
\bf Guanheng Chen$^{1}$~~~~~He Zhu$^{2}$~~~~~Xun Wu~~~~~Shaohan Huang~~~~~Furu Wei \\
~Microsoft Research \\
$^{1}$~Tsinghua University \\
$^{2}$~Peking University \\
~{\href{https://aka.ms/GeneralAI}{https://aka.ms/GeneralAI}}
}

\begin{document}

\maketitle

\footnotetext[1]{This is Part~III of Experiential Learning series. Part~I: \href{https://arxiv.org/abs/2602.12275}{On-Policy Context Distillation for Language Models}. Part~II: \href{https://arxiv.org/abs/2603.16856}{Online Experiential Learning for Language Models}.}

\begin{abstract}
Reinforcement learning (RL) on open-ended tasks compresses an LLM's rubric-based evaluation into a scalar reward, discarding rich textual feedback and conflating responses with distinct quality profiles.
We propose \textbf{Experiential Learning} (\ours{}), which repurposes the feedback model from an LLM-as-a-Judge into an LLM-as-a-Coach. The coach distills its assessment of each on-policy response into transferable experiential knowledge, which conditions a teacher model and is internalized by the policy through on-policy context distillation. Compared with scalar rewards, this higher-bandwidth feedback channel provides dense supervision and preserves fine-grained preferences among high-quality responses.
Across two policy families, with feedback from the policy itself or a proprietary model, \ours{} consistently outperforms rubric-based RL on held-out and unseen open-ended tasks. Notably, \ours{} generalizes better beyond the training distribution, and mitigates reward hacking. These findings establish experiential knowledge as a richer and more generalizable learning signal for post-training on non-verifiable tasks.
\begin{table}[H]
\centering
\begin{tabular}{@{}r@{\hspace{2pt}}l@{}}
\github & \textbf{Code}: \href{https://aka.ms/el-code}{\texttt{aka.ms/el-code}}
\end{tabular}
\end{table}
\end{abstract}

\section{Introduction}
\label{sec:intro}

\begin{figure}[ht]
\centering
\includegraphics[width=\linewidth]{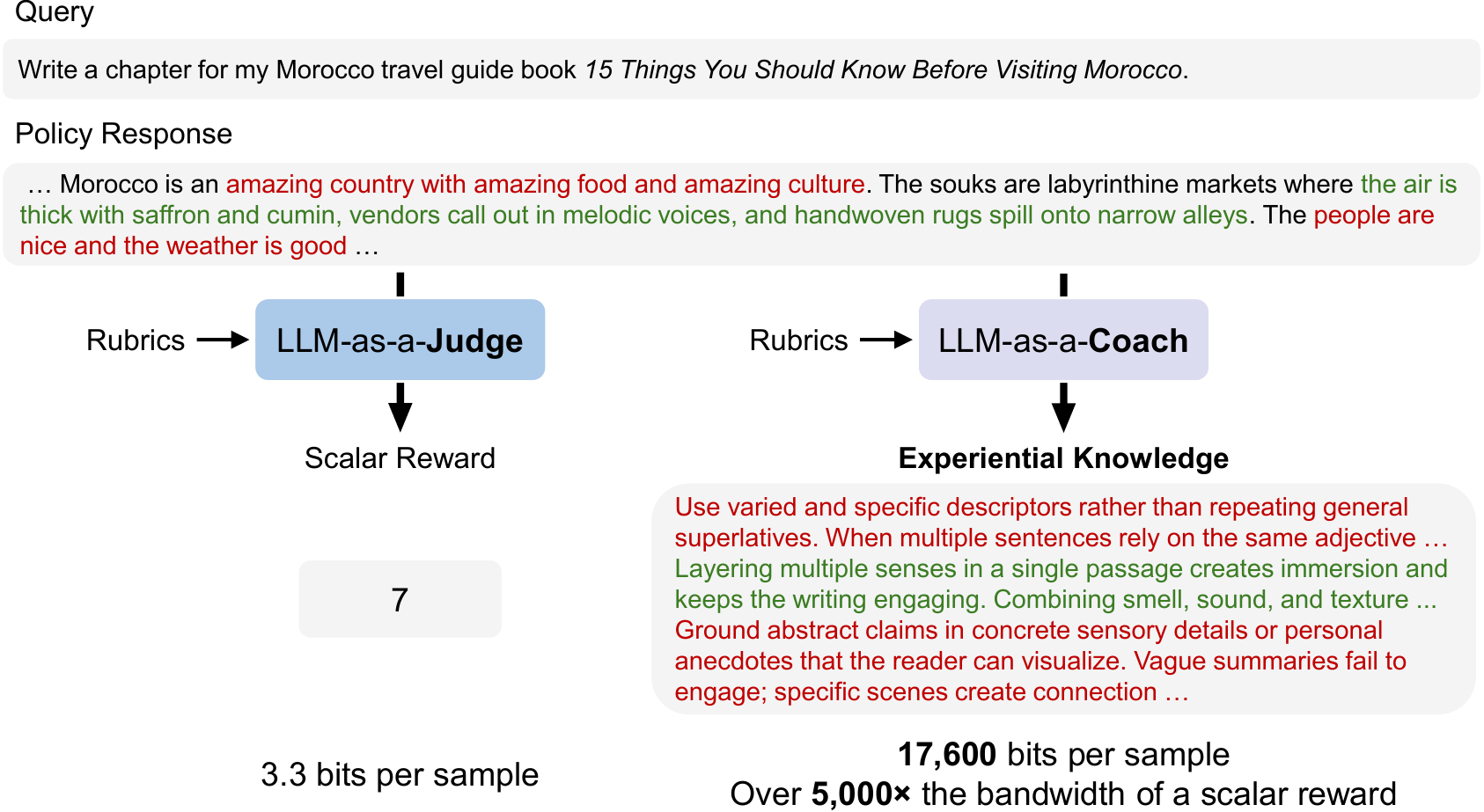}
\caption{Comparison of the learning signals in RL and \ours{}. RL compresses evaluation into a scalar reward, while \ours{} learns from experiential knowledge generated by an LLM-as-a-Coach. This richer signal speeds up learning and distinguishes top-performing responses that scalar rewards conflate on non-verifiable tasks. Red and green highlights connect weaknesses and strengths in the response to corresponding corrective and positive knowledge. RL bandwidth assumes a discrete 1--10 reward.}
\vspace{-0.1cm}
\label{fig:intro}
\end{figure}

Reinforcement learning~\citep{rlintro} has become a standard approach for post-training language models. It is particularly effective on verifiable tasks, where mathematical answers can be checked, programs can be executed, and constrained outputs can be validated automatically~\citep{tulu3}. Open-ended tasks, however, rarely have a unique correct answer. The quality of an explanation, recommendation, or creative response instead depends on multiple dimensions, such as factuality, relevance, completeness, organization, and style. Training on such tasks therefore often relies on an LLM-as-a-Judge that evaluates responses according to natural-language rubrics~\citep{mtbench,rubric-as-reward}.

Standard RL reduces this evaluation to a scalar reward. Although the evaluator may identify specific strengths, failures, and possible improvements, only the final score is used for optimization, while the remaining textual analysis is discarded. This creates an information bottleneck: responses receiving the same score become indistinguishable to the optimizer, even when the evaluator expresses meaningful preferences among them. The problem is especially pronounced near the top of the reward scale, where many satisfactory responses may receive the same maximum score despite differing in subtle but important ways.

In this work, we introduce \textbf{Experiential Learning} (\ours{}), which learns directly from the rich knowledge produced during evaluation. As illustrated in \Cref{fig:intro}, \ours{} repurposes the feedback model from an \textit{LLM-as-a-Judge} into an \textit{LLM-as-a-Coach}. Given a prompt, an on-policy response, and a set of rubrics, the coach analyzes the response and distills its assessment into \emph{experiential knowledge}: general and transferable guidance for producing better responses to similar tasks. Rather than preserving instance-specific corrections, this knowledge captures reusable strategies derived from the policy's own experience.

To internalize this knowledge, \ours{} uses on-policy context distillation~\citep{opcd}. The extracted experience is provided as context to a teacher model, inducing a distribution that reflects the coach's guidance. Along responses sampled from the policy, we then minimize the token-level reverse KL divergence between the policy and the context-conditioned teacher. This converts natural-language feedback into dense distributional supervision and consolidates it into the policy parameters, so neither the coach nor the experiential context is needed at inference time.

Experiential knowledge provides a potentially higher-bandwidth feedback channel than a scalar reward, as illustrated in \Cref{fig:intro}. A discrete 1--10 score has a theoretical maximum bandwidth of 3.3 bits per sample; a learned reward head producing a bfloat16 (bf16) value provides at most 16 representational bits. Under the same alphabet-size calculation, an experiential context of 1024 tokens drawn from a vocabulary of size 150,000 has a theoretical upper bound of 17,600 bits---over $1{,}000\times$ the maximum bandwidth of a scalar reward. This comparison is a feedback-bandwidth intuition rather than a measure of usable supervision.
This information is translated by the context-conditioned teacher into dense, per-token supervision, accelerating learning when on-policy samples are limited. In the well-converged regime, where RL has saturated the available reward signal, scalar feedback conflates high-quality responses assigned the same reward level. Experiential knowledge retains these fine-grained preferences, enabling \ours{} to further improve responses on non-verifiable tasks involving multiple quality dimensions and nuanced trade-offs.

We evaluate \ours{} on open-ended instruction-following tasks across two policy families, using either the initial frozen policy checkpoint or a proprietary model accessed through an API as the feedback model. Across a held-out in-distribution evaluation and multiple unseen benchmarks, \ours{} consistently improves over rubric-based RL, demonstrating the effectiveness of experiential supervision across different policy and feedback models.

Our results further suggest that scalar-reward optimization is susceptible to training-set reward overoptimization, a form of reward hacking~\citep{reward_hacking}, whereas the improvements from \ours{} transfer more effectively beyond the training distribution. A controlled distribution-matching analysis illustrates that, under its simplified setup, quantized scalar feedback preserves a binary target but discards finer distinctions in a multimodal target, whereas distributional guidance more faithfully recovers the target distribution. We also find that distilled experiential knowledge is more effective than using raw critiques or rubrics. Iterative teacher updates improve task performance, while on-policy distillation over general-domain prompts mitigates forgetting in out-of-distribution capabilities.

\section{Method}
\label{sec:method}

\begin{figure}[t]
\centering
\includegraphics[width=0.96\linewidth]{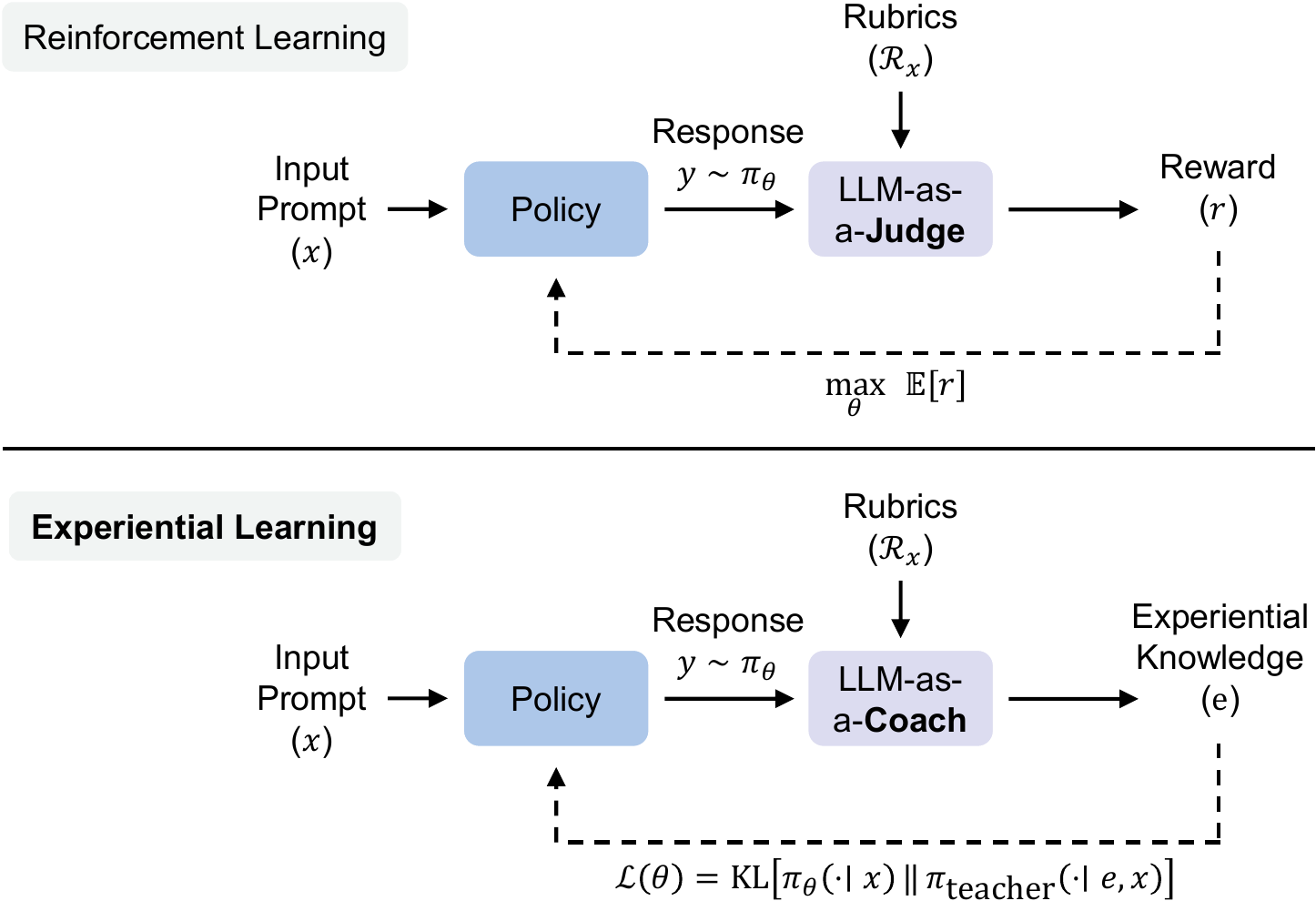}
\vspace{-0.1cm}
\caption{Comparison of RL and \ours{}. Reinforcement learning updates the policy by maximizing rewards assigned by an LLM-as-a-Judge, whereas experiential learning learns from experiential knowledge provided by an LLM-as-a-Coach under the same rubrics. Experiential knowledge provides richer learning signals for policy improvement than scalar rewards.}
\vspace{-0.1cm}
\label{fig:method}
\end{figure}

We present \textbf{Experiential Learning} (\ours{}) to address the problem of training a language model policy $\pi_\theta$ on non-verifiable open-ended tasks, where no ground-truth answer exists and response quality is assessed by another model.
Given a prompt $x$, the policy generates a response $y$. A feedback LLM $M$ then assesses the response against a set of rubrics $\mathcal{R}_x$ that decompose quality into checkable criteria. We denote the full output of $M$ as $M(x, y, \mathcal{R}_x)$, from which the policy is optimized.
In standard RL, $M$ acts as an LLM-as-a-Judge and reduces its assessment to a scalar reward, discarding all other textual content. In contrast, \ours{} uses $M$ as an LLM-as-a-Coach that extracts experiential knowledge to guide policy optimization, exposing a potentially higher-bandwidth signal than scalar feedback.

\subsection{Preliminaries: RL with an LLM-as-a-Judge}
Before introducing the learning objective of \ours{}, we formalize the standard RL baseline, in which the feedback LLM $M$ serves as an \textit{LLM-as-a-Judge}~\citep{mtbench,geval,genreward,reward-reason}. For each on-policy response $y \sim \pi_\theta(\cdot \mid x)$, only a scalar reward is extracted, and the remaining textual output is discarded. The policy is trained on a dataset $\mathcal{D}$ of prompts and rubrics to maximize the expected reward:
\begin{equation}
\begin{gathered}
\max_\theta\ \mathbb{E}_{(x, \mathcal{R}_x) \sim \mathcal{D},\, y \sim \pi_\theta(\cdot \mid x)} \left[ r \right], \\
\text{where } r = \mathrm{Extract\_Reward}(M(x, y, \mathcal{R}_x))
\end{gathered}
\end{equation}
We use GRPO~\citep{grpo} to implement the RL baseline.

\subsection{Experiential Learning with an LLM-as-a-Coach}
In the coach interface, \ours{} repurposes the same feedback LLM $M$ as an \textit{LLM-as-a-Coach}. For each on-policy response, $M$ analyzes the rubric-based assessment and distills transferable insights into experiential knowledge $e$. The policy is then trained to minimize the reverse KL divergence between its own distribution $\pi_\theta(\cdot \mid x)$ and a context-conditioned teacher distribution $\pi_\mathrm{teacher}(\cdot \mid e, x)$. The loss function is defined as:
\begin{equation}
\begin{gathered}
\mathcal{L}(\theta) = \mathbb{E}_{(x, \mathcal{R}_x) \sim \mathcal{D},\, y \sim \pi_\theta(\cdot \mid x)} \left[ \frac{1}{|y|} \sum_{t=1}^{|y|} D_\mathrm{KL}\left( \pi_\theta(\cdot \mid x, y_{<t}) \,\|\, \pi_\mathrm{teacher}(\cdot \mid e, x, y_{<t}) \right) \right], \\
\text{where } e = \mathrm{Extract\_Knowledge}(M(x, y, \mathcal{R}_x))
\end{gathered}
\end{equation}
The teacher can be either the initial frozen policy checkpoint or an iteratively updated version, where the policy checkpoint at the end of each epoch serves as the teacher for the next. \ours{} uses the experiential knowledge produced by the LLM-as-a-Coach as the learning signal, internalizing its rich distributional information into the policy weights via on-policy context distillation~\citep{opcd}. Refer to Appendix~\ref{app:exp_detail_templates} for prompt templates.

\begin{algorithm}[t]
\small
\caption{\ours{}: Experiential Learning}
\label{alg:ours}
\begin{algorithmic}
\Require Training data $\mathcal{D} = \{(x, \mathcal{R}_x)\}$; Policy $\pi_\theta$; Teacher $\pi_\mathrm{teacher}$; LLM-as-a-Coach $M$
\Ensure Trained policy $\pi_\theta$
\For{each batch $x \sim \mathcal{D}$}
    \State \textcolor{gray}{\textit{// On-policy rollout}}
    \State Sample responses $y \sim \pi_\theta(\cdot \mid x)$
    \Statex
    \State \textcolor{gray}{\textit{// Extract experiential knowledge}}
    \State $e \gets \mathrm{Extract\_Knowledge}(M(x, y, \mathcal{R}_x))$
    \Statex
    \State \textcolor{gray}{\textit{// Compute token-level reverse KL toward context-conditioned teacher}}
    \State $\mathcal{L}(\theta) \gets \frac{1}{|y|} \sum_{t=1}^{|y|} D_\mathrm{KL}\!\left( \pi_\theta(\cdot \mid x, y_{<t}) \,\|\, \pi_\mathrm{teacher}(\cdot \mid e, x, y_{<t}) \right)$
    \Statex
    \State \textcolor{gray}{\textit{// Update policy}}
    \State Update $\theta$ by minimizing $\mathcal{L}(\theta)$
\EndFor
\State \Return $\pi_\theta$
\end{algorithmic}
\end{algorithm}

\subsection{From Judge to Coach}
The same feedback LLM $M$ and prompt-specific rubrics $\mathcal{R}_x$ support both learning interfaces, but the two interfaces use the resulting assessment differently, as summarized in \Cref{tab:judge-coach}.

\begin{table}[ht]
\centering
\small
\resizebox{0.9\linewidth}{!}{
\begin{tabular}{@{}lll@{}}
\toprule
 & LLM-as-a-Judge & \bf LLM-as-a-Coach (Ours) \\
\midrule
Role & Evaluate: assess response quality & Guide: extract transferable guidance \\
Learning Signal & Sequence-level scalar score & Experiential knowledge \\
Feedback Bandwidth & Low & High \\
\bottomrule
\end{tabular}
}
\vspace{0.15cm}
\caption{Comparison of an LLM-as-a-Judge and an LLM-as-a-Coach in their roles, learning signals, and feedback bandwidths. Both use the same feedback LLM and rubrics.}
\label{tab:judge-coach}
\end{table}

An LLM-as-a-Judge evaluates a response and exposes only a scalar score as the learning signal. An LLM-as-a-Coach instead guides improvement through experiential knowledge. Put simply, a judge tells the policy how well it performed, whereas a coach provides actionable guidance for performing better.

Importantly, this distinction concerns the feedback consumed by the optimizer, not the intrinsic capabilities of $M$. An LLM-as-a-Judge may produce detailed textual analysis, but standard RL retains only its final score, creating a low-bandwidth learning channel. In \ours{}, the LLM-as-a-Coach distills the same rubric-based assessment into transferable experiential knowledge. A context-conditioned teacher then translates this higher-bandwidth feedback into dense token-level supervision. We develop this bandwidth intuition below.

\subsection{Feedback-Bandwidth Intuition}

A useful intuition for the advantage of \ours{} over RL comes from the maximum representational bandwidth of their respective feedback channels.

\paragraph{RL} In our experiments, we adopt a discrete 1--10 scoring scale following standard rubric-based evaluation practice, whose theoretical maximum bandwidth is $\log_2(10) \approx 3.3$ bits per sample. If rewards are produced in bfloat16 (bf16) using a learned projection head, the maximum representational bandwidth increases to $16$ bits.

\paragraph{\ours{}} The teacher $\pi_\mathrm{teacher}$ is either the initial frozen policy checkpoint or an iteratively updated policy checkpoint. Together, $\pi_\theta$ and $\pi_\mathrm{teacher}$ form a closed system whose external learning signal is the experiential knowledge $e$ produced by $M$.
A theoretical upper bound on the bandwidth of this textual channel is determined by $e$: a context of $L$ tokens drawn from a vocabulary of size $|\mathcal{V}|$ yields up to $L \cdot \log_2(|\mathcal{V}|)$ bits.\footnote{This quantity is the theoretical maximum encoding capacity of the textual channel. It does not measure the effective supervision carried by the experiential knowledge, nor its mutual information with the desired policy improvement; natural-language redundancy and imperfect knowledge extraction can make the usable information substantially smaller.} With $L{=}1024$ and $|\mathcal{V}|{=}150{,}000$, this amounts to approximately $1024 \times 17.2 \approx 17{,}600$ bits per sample---over \textbf{1,000$\times$} the theoretical maximum bandwidth of a bf16 scalar reward and \textbf{5,000$\times$} of a discrete 1--10 reward.

\textbf{This potentially wider feedback channel can be advantageous on non-verifiable tasks.} For verifiable tasks~\citep{tulu3} with a binary correctness objective, a single bit suffices to represent the reward outcome. However, non-verifiable tasks involve multiple quality dimensions, stylistic preferences, and nuanced trade-offs. In the converged regime where rollouts are abundant and RL has saturated the reward signal, all responses achieving the maximum reward level become \textbf{indistinguishable} due to the information bottleneck, limiting further alignment with the fine-grained preferences expressed by the LLM-as-a-Judge.
\ours{} bypasses this bottleneck by transmitting richer information through the experiential knowledge context, enabling a \textbf{higher-fidelity approximation of the target distribution defined by $M$}.

\section{Experiments}

For non-verifiable tasks, on-policy training with rubrics provides a principled way to assess open-ended responses from the policy model. Standard RL uses an LLM-as-a-Judge to evaluate each response against rubrics, producing a scalar reward while discarding the textual feedback. \ours{} instead employs an LLM-as-a-Coach that analyzes each response against the same rubrics and generates experiential knowledge, which is then consolidated into the policy model weights.

\subsection{Setup}

\paragraph{Training Data and Models}
We source training prompts from WildChat-IF~\citep{wildchat,golf,rlmt}, a curated collection of 7500 real user queries emphasizing diverse conversational instructions. For each prompt, we pre-generate a set of evaluation rubrics using GPT-4o. These rubrics decompose response quality into fine-grained, independently checkable criteria. In experiments with iterative teacher model, we mix in prompts from Tulu3~\citep{tulu3} at a 1:0.25 ratio to preserve general capabilities. These prompts use the initial frozen policy checkpoint as the OPD teacher and carry no extra context. We filter the Tulu3 SFT mixture to exclude WildChat-overlapping sources.
We use Qwen3-8B~\citep{qwen3} (non-thinking mode) and OLMo-3-7B-Instruct~\citep{olmo} as the policy models. We use either the initial frozen policy checkpoint or GPT-4o as the LLM-as-a-Judge for RL and the LLM-as-a-Coach for \ours{}.

\paragraph{Training}
We use Rubric-as-Reward~\citep{rubric-as-reward} implemented with GRPO~\citep{grpo} as the baseline, which optimizes the policy using a 1--10 rating from the LLM-as-a-Judge as the scalar reward. For \ours{}, the LLM-as-a-Coach generates experiential knowledge based on the rubrics and the response, which is then prepended as context to the teacher model. The teacher can be the initial frozen policy checkpoint or an iteratively updated version where the policy checkpoint at the end of each epoch becomes the teacher for the next epoch.
Both methods use a batch size of 256 prompts and 8 sampled responses per prompt. Learning rate is set to 1e-6. The maximum prompt length is 8192 tokens, and the maximum response length is 4096 tokens. Rollout temperature is set to 1. Training runs for 3 epochs over the dataset and checkpoints are saved every 10 steps. Reverse KL is computed over the top 256 vocabulary tokens ranked by the student model's predicted probabilities without re-normalization to 1. Refer to Appendix~\ref{app:exp_detail_templates}--\ref{app:e2e_example} for prompt templates and examples.

\paragraph{Evaluation}
We evaluate on a held-out test set of 250 WildChat-IF prompts using GPT-4o as a rubric-conditioned LLM-as-a-Judge, sampling 4 responses per prompt.
We also measure transfer to four open-ended benchmarks not seen during training: AlpacaEval~v2.0~\citep{alpacaeval}, WildBench~\citep{wildbench}, ArenaHard~v2.0~\citep{arenahard}, and CreativeWritingV3~\citep{creativewritingv3}. GPT-4o serves as the evaluator across all benchmarks. The first three are pairwise evaluations against GPT-4-Turbo references: we report win rate (\%) for AlpacaEval~v2.0 and ArenaHard~v2.0, and a preference score in $[-100, 100]$ for WildBench. CreativeWritingV3 uses direct rubric-based scoring, reported in $[0, 100]$.
We additionally cross-checked the main conclusions with a stronger proprietary evaluator and observed consistent relative trends. We therefore report GPT-4o evaluations throughout to reduce evaluation cost.

\subsection{Main Results}

\begin{table}[t]
\centering
\small
\begin{tabular}{@{}llccccc}
\toprule
\multirow{2.5}{*}{\tabincell{l}{\textbf{Policy +} \\ \textbf{Feedback}}} & \multirow{2.5}{*}{\textbf{Method}} & \textbf{WildChat} & \textbf{AlpacaEval-v2} & \textbf{WildBench} & \textbf{ArenaHard-v2} & \textbf{CW-v3} \\
\cmidrule(lr){3-3}
\cmidrule(lr){4-4}
\cmidrule(lr){5-5}
\cmidrule(lr){6-6}
\cmidrule(lr){7-7}
& & Score & Win Rate (\%) & Score & Win Rate (\%) & Score \\
\midrule
\multirow{3}{*}{\tabincell{l}{Qwen3-8B \\ + Qwen3-8B}} & Base Model & \textcolor{gray}{78.1} & \textcolor{gray}{34.5} & \textcolor{gray}{17.6} & \textcolor{gray}{29.1} & \textcolor{gray}{73.8} \\
& RL & 79.2 & 37.3 & 18.4 & 30.5 & 74.3 \\
& \textbf{\ours{}} & \textbf{80.0} & \textbf{40.0} & \textbf{21.8} & \textbf{31.0} & \textbf{76.0} \\
\midrule
\multirow{2}{*}{\tabincell{l}{Qwen3-8B \\ + GPT-4o}} & RL & 80.4 & \textbf{38.2} & 19.2 & 31.2 & 74.4 \\
& \textbf{\ours{}} & \textbf{80.7} & 37.4 & \textbf{22.0} & \textbf{31.9} & \textbf{75.3} \\
\midrule\midrule
\multirow{3}{*}{\tabincell{l}{OLMo-3-7B \\ + OLMo-3-7B}} & Base Model & \textcolor{gray}{75.7} & \textcolor{gray}{43.6} & \textcolor{gray}{39.0} & \textcolor{gray}{21.8} & \textcolor{gray}{78.7} \\
& RL & 76.0 & 45.9 & 42.1 & 23.3 & \textbf{78.8} \\
& \textbf{\ours{}} & \textbf{77.1} & \textbf{50.8} & \textbf{47.5} & \textbf{26.2} & \textbf{78.8} \\
\midrule
\multirow{2}{*}{\tabincell{l}{OLMo-3-7B \\ + GPT-4o}} & RL & 76.8 & 44.7 & 40.3 & 22.5 & 77.8 \\
& \textbf{\ours{}} & \textbf{78.2} & \textbf{48.5} & \textbf{46.2} & \textbf{25.1} & \textbf{78.0} \\
\bottomrule
\end{tabular}
\vspace{0.2cm}
\caption{Evaluation scores of RL and \ours{}. ``Policy + Feedback'' denotes the policy model paired with the feedback model, which acts as an LLM-as-a-Judge for RL or an LLM-as-a-Coach for \ours{}. CW-v3 is short for CreativeWritingV3. \ours{} outperforms RL on most benchmarks across all model combinations.}
\label{tab:main-results}
\end{table}

We present the main results in \Cref{tab:main-results}. In the ``Policy + Feedback'' column, the first model is the policy being optimized, and the second is the feedback LLM, acting as an LLM-as-a-Judge for RL and an LLM-as-a-Coach for \ours{}. We consider two choices of feedback LLM: the initial frozen policy checkpoint and the closed-source GPT-4o. We report the average score (or win rate) across the top-3 performing checkpoints.

Both RL and \ours{} are trained on WildChat-IF prompts and evaluated on the held-out WildChat test set together with four additional benchmarks. \ours{} consistently outperforms RL on the WildChat evaluation and generalizes better to unseen benchmarks, with particularly large gains on AlpacaEval v2.0 and WildBench. This advantage holds whether the feedback LLM is the initial frozen policy checkpoint or the closed-source GPT-4o, indicating that the benefit of \ours{} stems from its richer information channel rather than from a specific feedback LLM. We further observe that using GPT-4o as the feedback LLM improves WildChat scores for both RL and \ours{} relative to using the initial frozen policy checkpoint.

\subsection{Experiential Learning Generalizes Better}

\begin{wrapfigure}{r}{8cm}
\centering
\vspace{-0.45cm}
\includegraphics[width=0.58\textwidth]{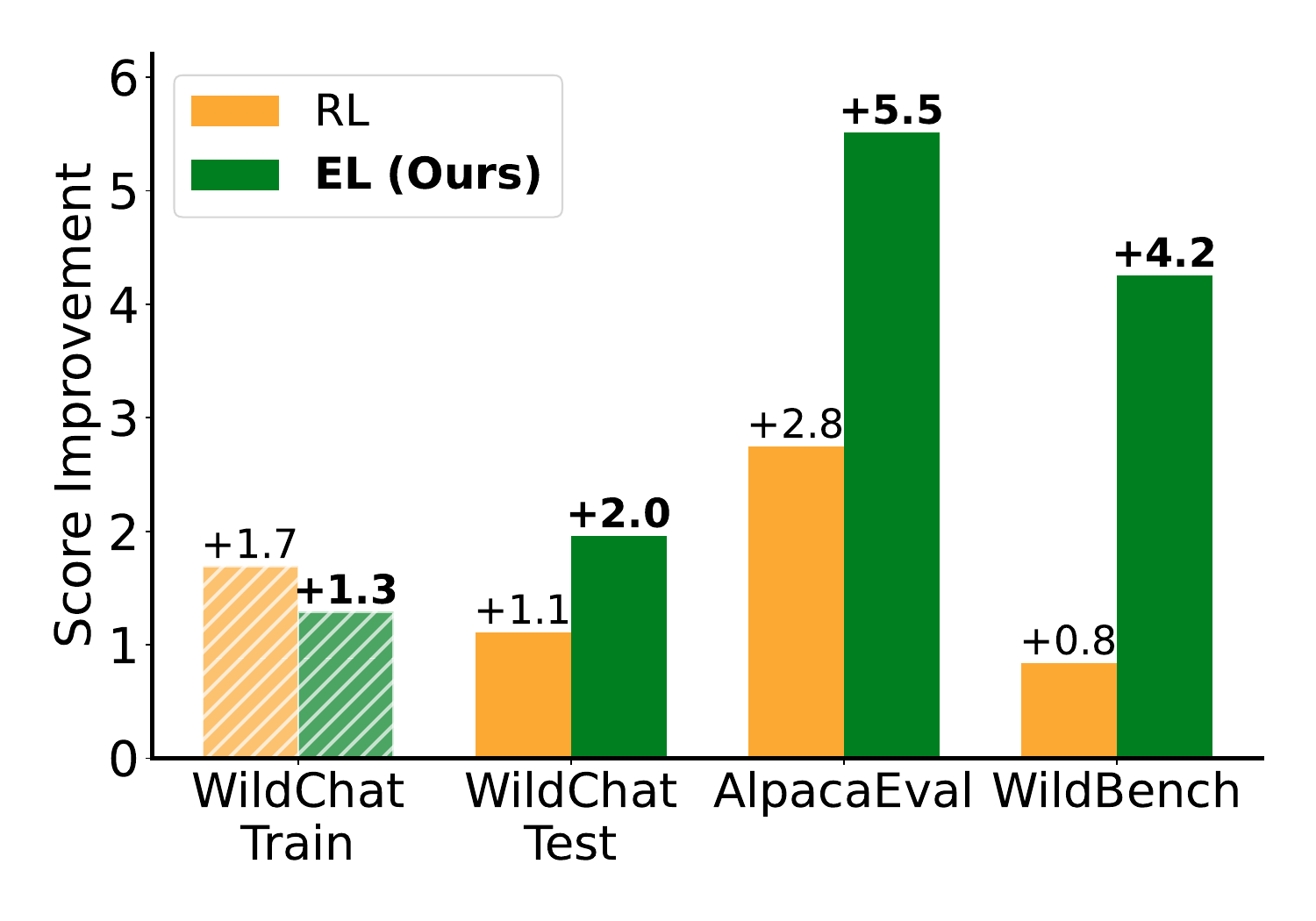}
\vspace{-0.45cm}
\caption{Score improvement over the base model. ``WildChat Train'' is evaluated on a subset of the training data after training completes. \ours{} generalizes better than RL despite achieving lower training-set scores.}
\vspace{-0.1cm}
\label{fig:generalization}
\end{wrapfigure}

We sample a subset of 1000 WildChat training examples and evaluate on it after training completes. We compare the results with those on WildChat test set and two benchmarks, as shown in \Cref{fig:generalization}. Although \ours{} achieves a smaller gain than RL on the training set, it obtains higher scores on the WildChat test set, AlpacaEval~V2.0, and WildBench.

We hypothesize that this difference arises in part from the feedback-bandwidth difference between RL and \ours{}. RL maximizes a scalar reward that carries limited information, allowing the policy to drift toward patterns that inflate training rewards without transferring---a manifestation of reward hacking~\citep{reward_hacking}. \ours{} never receives a scalar score from the LLM-as-a-Coach during training. Instead, its learning signal is the full distributional shift induced by the experiential knowledge context, which steers the policy toward more transferable behaviors. We use Qwen3-8B as both the policy model and the feedback LLM in \Cref{fig:generalization}. We observe a similar phenomenon when replacing the Qwen3-8B feedback LLM with GPT-4o.

\subsection{Ablations}

\paragraph{Iterative Teacher}

\begin{wrapfigure}{r}{6.7cm}
\centering
\resizebox{0.40\textwidth}{!}{
\begin{tabular}{@{}lcc@{}}
\toprule
\bf Configuration & \bf Score & \bf OOD Acc. \\
\midrule
Base Model & \textcolor{gray}{78.1} & \textcolor{gray}{83.5} \\
RL & 79.2 & 81.3 \\
\midrule
Fixed Teacher & 80.0 & 83.1 \\
Iterative & 80.7 & 74.9 \\
Iterative+General & 80.7 & 79.9 \\
\bottomrule
\end{tabular}
}
\makeatletter
\def\@captype{table}
\makeatother
\caption{Ablation on iterative teacher updates. ``Score'' is the WildChat test set score averaged over the top-3 performing checkpoints and ``OOD Acc.'' is the IFEval accuracy averaged over the same checkpoints.}
\label{tab:iterative-teacher}
\vspace{-1em}
\end{wrapfigure}

We ablate the teacher update strategy in \Cref{tab:iterative-teacher}, using Qwen3-8B as both the policy model and the LLM-as-a-Coach.
In the \textit{Fixed Teacher} setting, the teacher remains the initial frozen policy checkpoint throughout training, which is also the default setting of our experiments.
In the \textit{Iterative} setting, the policy checkpoint at the end of each epoch becomes the teacher for the next epoch.
In the \textit{Iterative+General} setting, we mix Tulu3~\citep{tulu3} prompts into WildChat-IF at a 1:0.25 ratio. These general prompts carry no experiential context and use the initial frozen policy checkpoint as the OPD teacher~\citep{mopd,thinkingmachine-onpolicy}; we filter the Tulu3 SFT mixture to exclude WildChat-overlapping sources.

Iteratively updating the teacher improves the WildChat score from 80.0 to 80.7, but substantially reduces IFEval accuracy, indicating forgetting on out-of-distribution instruction following. Adding general-prompt OPD from the initial frozen policy checkpoint recovers much of this loss, improving IFEval from 74.9 to 79.9 while preserving the same WildChat score. This makes general-prompt OPD a useful stabilizer when applying iterative teacher updates.

\begin{wrapfigure}{r}{6.7cm}
\centering
\resizebox{0.40\textwidth}{!}{
\begin{tabular}{@{}lcc@{}}
\toprule
\bf Configuration & \bf Score & \bf OOD Acc. \\
\midrule
Base Model & \textcolor{gray}{64.8} & \textcolor{gray}{68.0} \\
RL & 67.7 & 67.6 \\
\midrule
Full Critique & 67.9 & 64.6 \\
Rubrics Only & 68.1 & 65.6 \\
Multiple-Choice & 66.3 & 68.5 \\
\textbf{Default \ours{}} & \textbf{68.6} & \textbf{68.8} \\
\bottomrule
\end{tabular}
}
\makeatletter
\def\@captype{table}
\makeatother
\caption{Ablation of teacher context on Qwen3-1.7B. ``Score'' is WildChat test set score (Top-3 Average); ``OOD Acc.'' is the IFEval accuracy averaged over the same three checkpoints.}
\label{tab:ablations}
\vspace{-1em}
\end{wrapfigure}

\paragraph{Teacher Context}
We ablate the format of context provided by $M$ in \Cref{tab:ablations}, using Qwen3-1.7B as the policy model and Qwen3-4B as the LLM-as-a-Coach. ``Score'' denotes the WildChat test set score evaluated by GPT-4o and averaged over the top-3 performing checkpoints. ``OOD Acc.'' denotes the IFEval~\citep{ifeval} accuracy for evaluating out-of-distribution capability, averaged over the same three checkpoints.

\textit{Full Critique} prepends the entire scoring output from $M$, including the per-rubric analysis, as context for the teacher, without performing experiential knowledge extraction.
\textit{Rubrics Only} provides the rubrics as context without any response-specific feedback~\citep{rubric-sd,rcsd}.
\textit{Multiple-Choice} asks the LLM-as-a-Coach to select one of 10 predefined improvement directives (e.g., ``Improve factual accuracy''), compressing feedback to a single categorical choice. The maximum feedback bandwidth of this configuration reduces to $\log_2 10 \approx 3.3$ bits, equivalent to that of a 1--10 scalar reward in RL.
We also experimented with including the response from the policy model alongside the full scoring output in the context. This configuration diverges within a few training steps.
Prompt templates for all configurations are provided in Appendix~\ref{app:ablation_templates}.

All variants improve over the base model on the WildChat test set, and the default \ours{} setting that extracts transferable experiential knowledge performs best. \textit{Full Critique} and \textit{Rubrics Only} degrade IFEval accuracy because evaluation-oriented context biases the teacher toward a critiquing distribution rather than a task-solving distribution. The policy then imitates this misaligned behavior, leading to forgetting on OOD tasks. \textit{Multiple-Choice} preserves OOD performance but offers limited gains, consistent with its lower feedback bandwidth.

\subsection{Analysis}

\paragraph{Feedback Bandwidth Comparison}

\Cref{tab:info_capacity} gives a qualitative comparison of the feedback bandwidth of three learning methods. RL receives a scalar reward per sample, with $O(1)$ representational bandwidth regardless of response length. On-policy distillation (OPD) from a stronger teacher can provide feedback that scales as $O(N)$, where $N$ is the response length, since the teacher's per-token distribution encodes knowledge from its larger parameter space. In \ours{}, the student and the teacher form a closed system. The external learning signal comes entirely from the experiential knowledge context of length $L$ produced by the coach, so its theoretical maximum textual bandwidth scales as $O(L)$. Although both OPD and \ours{} use distillation, their information sources differ: OPD draws from the teacher's superior parametric knowledge, whereas \ours{} draws from the coach's rubric-based assessment externalized as text.

For RL, a discrete 1--10 score has a theoretical maximum bandwidth of 3.3 bits per sample. A learned reward head that produces a BF16 value has a maximum representational bandwidth of 16 bits, but the reward model uses only a small subspace of this range in practice.
The sparse signal landscape can make it easier for the policy to exploit by reward hacking rather than providing more usable information.

\begin{table}[ht]
\centering
\small
\resizebox{\linewidth}{!}{
\begin{tabular}{@{}llc@{}}
\toprule
\bf Method & \bf Feedback Source & \bf Bandwidth \\
\midrule
Reinforcement Learning (RL) & Sequence-level scalar reward from LLM-as-a-judge & $O(1)$ \\
On-Policy Distillation (OPD) & Larger teacher parameters & $O(N)$ \\
On-Policy Self-Distillation (OPSD) & Privileged information, e.g., ground-truth answers, and hints    & $O(L)$ \\
\textbf{Experiential Learning (\ours{})} & Experiential knowledge context from LLM-as-a-coach & $O(L)$ \\
\bottomrule
\end{tabular}
}
\vspace{0.15cm}
\caption{Qualitative comparison of feedback bandwidth across learning methods. $N$: response length; $L$: experiential context length.}
\label{tab:info_capacity}
\end{table}

\paragraph{Distribution-Matching Analysis of Feedback Quantization}

\begin{figure}
\centering
\includegraphics[width=0.9\linewidth]{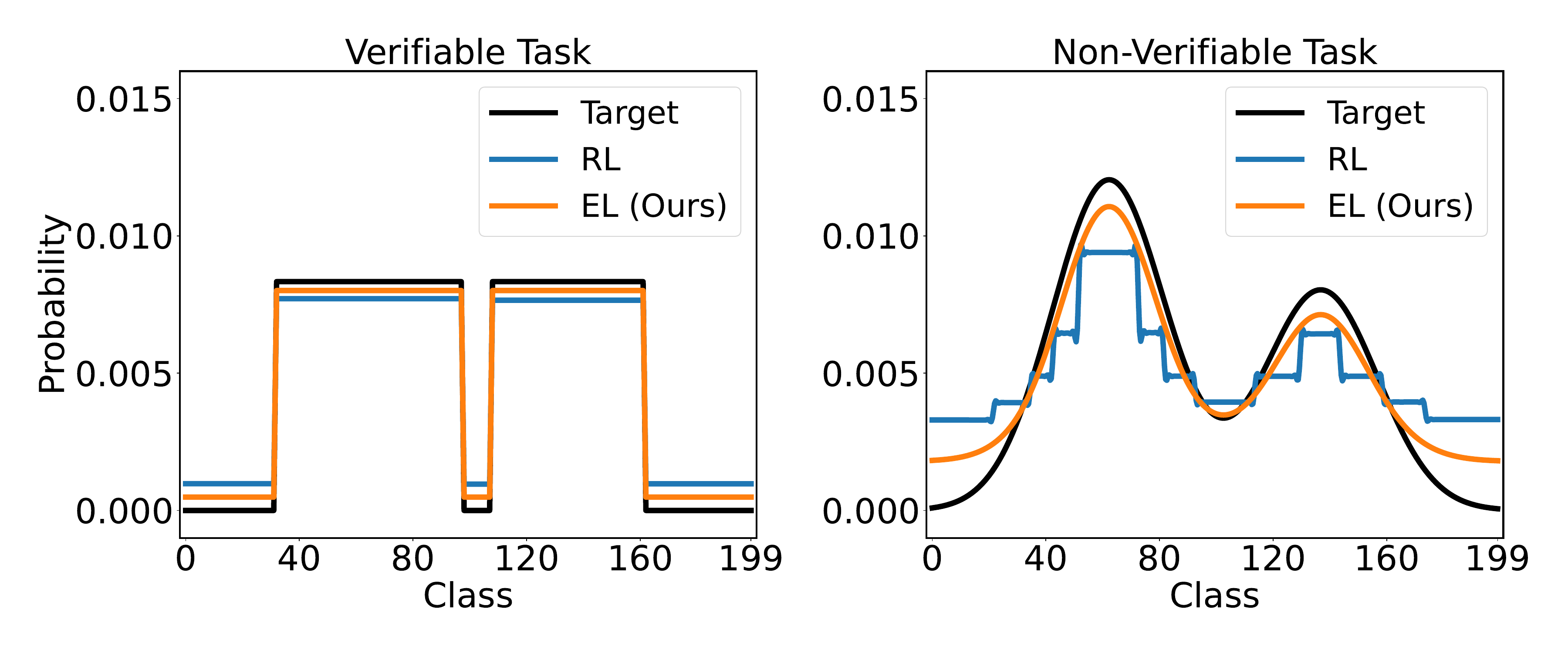}
\vspace{-0.15cm}
\caption{Controlled distribution-matching analysis. Left: for a binary target, RL with binary rewards and direct distributional matching converge to similar distributions. Right: for a bimodal target, the constructed five-level reward produces a staircase distribution, while direct distributional matching recovers the smooth target shape.}
\vspace{-0.1cm}
\label{fig:toy}
\end{figure}

We use a controlled toy setting to analyze how quantizing feedback into scalar levels can affect distribution matching. This construction isolates the representation of feedback: it is not intended as a faithful simulation of the full RL and \ours{} training pipelines, nor as a proof of a fundamental limitation shared by all scalar-reward methods. Rather, it provides intuition for the information discarded when a fine-grained target is mapped to a small number of reward levels. The results are shown in \Cref{fig:toy}.

Suppose the feedback LLM, acting as either an LLM-as-a-Judge or an LLM-as-a-Coach, has an internal ideal distribution $p^*$ over a discrete vocabulary of $V{=}200$ tokens that it wishes the policy to learn. We train a categorical policy $\pi_\theta = \text{softmax}(\theta)$ to approximate $p^*$ via on-policy sampling: at each step, a token $y$ is sampled from $\pi_\theta$ and a training signal is received. We compare two feedback mechanisms:
(1) \textbf{RL}: the policy receives a scalar reward quantized to $L$ levels: $r(y) = \left\lfloor \frac{p^*(y)}{\max_{y'} p^*(y')} \cdot (L-1) + 0.5 \right\rfloor, r(y) \in \{0, 1, \ldots, L{-}1\}$
and is optimized via REINFORCE~\citep{reinforce} with a KL penalty toward the uniform distribution.
(2) \textbf{Distributional guidance (an idealized proxy for \ours{})}: assuming that the experiential knowledge carries sufficient information to specify the target distribution, we directly minimize the reverse KL divergence $\text{KL}(\pi_\theta \| p^*)$. We consider two scenarios: a \textit{verifiable} task where $p^*$ is binary (correct/incorrect tokens, $L{=}2$), and a \textit{non-verifiable} task where $p^*$ is a bimodal Gaussian mixture quantized into $L{=}5$ reward levels.

As shown in \Cref{fig:toy} (left), on the constructed binary task both RL and direct distributional matching converge to nearly identical distributions matching $p^*$. Under this target and objective, responses are either correct or incorrect, and no finer-grained distinction is required; binary rewards therefore preserve the information needed for the specified target.

However, under the five-level reward construction (\Cref{fig:toy}, right), a clear gap emerges. Direct distributional matching recovers the smooth bimodal shape of $p^*$, while RL converges to a staircase-shaped distribution. Specifically, tokens at the peak plateau that receive the same maximum reward learn identical probability under this RL objective, even though $p^*$ assigns different likelihoods to them.
Within this setup, the quantized scalar reward cannot further distinguish among tokens assigned the same top score. Direct distributional guidance retains those fine-grained distinctions and therefore matches $p^*$ without the quantization artifacts.

\section{Discussion}

\paragraph{Reward Hacking}
\ours{} mitigates reward hacking that arises from the information bottleneck between the feedback model and the policy: when an LLM-as-a-Judge compresses its assessment into a scalar reward, the policy can exploit patterns that inflate the score without genuinely improving response quality. By transmitting richer information through experiential knowledge, \ours{} reduces this failure mode. We note that a separate source of reward hacking can occur when the feedback model itself is biased or miscalibrated. Since \ours{} addresses the \textit{feedback bandwidth} between the feedback model and the policy rather than the quality of the feedback model itself, improving the calibration and accuracy of the feedback model remains a complementary direction.

\paragraph{Distinction from Reinforcement Learning}
Although the reverse KL objective of \ours{} can be decomposed into a token-level advantage-weighted form, it differs from RL in its optimization target. RL maximizes a scalar reward: its optimal policy concentrates probability mass on the highest-reward responses. \ours{}, through on-policy context distillation, instead minimizes divergence from a context-conditioned teacher distribution. RL is therefore reward-seeking, while \ours{} is distribution-matching. The former exploits any feature correlated with higher reward, whereas the latter stays close to the teacher's distributional shape.

\paragraph{Training Cost and Efficiency}
Both RL and \ours{} spend most of their end-to-end training time in the online data-generation stage: sampling rollouts from the policy and invoking the feedback LLM as an LLM-as-a-Judge or LLM-as-a-Coach. These two expensive components are shared by both methods. \ours{} additionally requires a context-conditioned teacher forward pass and token-level distillation, whereas RL uses reward-based GRPO updates. In our setting, the cost of these method-specific updates is small relative to the shared rollout and feedback-generation cost.

\section{Related Work}

\paragraph{Learning from Experience}
Recent work advocates learning from experience through interactions with the environment~\citep{eraexp}. Such experience has been shown to accelerate future learning~\citep{earlyexp} and enable reasoning agents to discover effective strategies through self-play and reflection~\citep{tencentgame}. For language models, prior work leverages interaction histories by reflecting on past failures~\citep{reflexion}, storing reusable knowledge in external memory~\citep{expel}, or optimizing with multi-turn textual feedback~\citep{textgrad,gepa,feedback-descent}.
More recently, critiques have been incorporated into RL objectives by conditioning policy optimization on critique contexts~\citep{critique-grpo,golf}. However, these methods remain off-policy with respect to the original prompt, and the learning signal is ultimately bottlenecked by scalar rewards.

\paragraph{Context Distillation and Self-Distillation}
Context distillation aims to compress knowledge provided in context into model parameters, eliminating inference-time context processing~\citep{context:distill:anthropic,context:distill:berkeley,infiniteicl}.
To mitigate the exposure bias of off-policy training, on-policy distillation trains the student on its own generated trajectories~\citep{minillm,thinkingmachine-onpolicy,googlepolicy}.
Building on these ideas, on-policy context distillation (OPCD) performs context distillation with a context-conditioned teacher under on-policy training~\citep{opcd}.
A closely related line of work is on-policy self-distillation, where the teacher is similarly conditioned on context (e.g., solutions, demonstrations, or environmental feedback), but is typically instantiated by the student model itself~\citep{self:distill:meta,self:distill:eth,self:distill:mit,self:distill:mila}.
Recent advances have extended this paradigm to combining RLVR~\citep{sdvr,unify-grsd,rebellious-stu}, integrated it into agentic tasks~\citep{skillsd,tcod,opid}, introduced more improved training algorithms~\citep{paced,sod,opsdl,context-return,crisp,gopd}, and adapted it for online learning~\citep{oel,sdpo-online,openclaw-rl}. Concurrently, a growing body of work has investigated the mechanisms underlying both its empirical success and its potential degradation~\citep{rethink-opd,sd-degrade,embarrassing-sd}.

\section{Conclusion}

We introduced \textbf{Experiential Learning} (\ours{}), a framework for post-training language models on non-verifiable tasks using rich experiential knowledge rather than scalar rewards. By repurposing the feedback model as an LLM-as-a-Coach, \ours{} distills rubric-based assessments into transferable guidance and internalizes it through on-policy context distillation. This higher-bandwidth feedback channel provides dense supervision and preserves fine-grained preferences that scalar rewards discard. Across different policy families and feedback models, \ours{} consistently outperforms rubric-based RL on held-out and unseen open-ended tasks while exhibiting stronger generalization beyond the training distribution. Our findings highlight experiential knowledge as an informative and transferable learning signal, opening directions for scalable iterative learning, improved knowledge extraction, and broader post-training of language models on complex open-ended tasks.



\bibliographystyle{alpha}
\bibliography{opcd}

\newpage
\appendix

\section{Details of Experiments}
\label{app:exp_detail}

\subsection{Prompt Templates}
\label{app:exp_detail_templates}

For \ours{}, the LLM-as-a-Coach $M$ receives the same inputs but is additionally asked to distill transferable experiential knowledge. We use the prompt template in \Cref{fig:el_extract_prompt}. The experiential knowledge is extracted from the \texttt{<experience>} tags.

\begin{figure}[h]
    \begin{tcolorbox}
    You are an expert evaluator. Given a user prompt, a generated response, and a list of quality rubrics, please rate the overall quality of the response on a scale of 1 to 10 based on how well it satisfies the rubrics. \\
    Consider all rubrics holistically when determining your score. A response that violates multiple rubrics should receive a lower score, while a response that satisfies all rubrics should receive a higher score. \\ \\
    <prompt> \\
    \{instruction\} \\
    </prompt> \\ \\
    <response> \\
    \{response\} \\
    </response> \\ \\
    <rubrics> \\
    \{rubric\_list\_string\} \\
    </rubrics> \\ \\
    First, analyze the response against each rubric item, discussing how well the response meets or fails each criterion. Then, provide your final score as an integer between 1 and 10, wrapped in <score> and </score> tags. \\ \\
    After the score, distill your analysis into transferable experiential knowledge which is general, high-level, widely applicable insights that would help improve future responses to similar tasks. Focus on reusable strategies and patterns rather than details specific to this particular response. Output this knowledge wrapped in <experience> and </experience> tags. \\ \\
    Example ending: \\
    <score> your\_integer\_score\_from\_1\_to\_10 </score> \\ \\
    <experience> \\
    some experiential knowledge... \\
    </experience> \\ \\
    Your evaluation:
    \end{tcolorbox}
    \caption{Prompt template for \ours{} knowledge extraction. The text within the \texttt{<experience>} tags is extracted as experiential knowledge $e$.}
    \label{fig:el_extract_prompt}
\end{figure}

The extracted experiential knowledge is then prepended as context to the teacher model using the template in \Cref{fig:el_context_prompt}.

\begin{figure}[h]
    \begin{tcolorbox}
    Here is some experiential knowledge: \\ \\
    <experience> \\
    \{experience\} \\
    </experience> \\ \\
    Given the above experiential knowledge, solve the following problem: \\
    \{prompt\}
    \end{tcolorbox}
    \caption{Prompt template for conditioning the teacher model on experiential knowledge.}
    \label{fig:el_context_prompt}
\end{figure}

For RL, the LLM-as-a-Judge $M$ evaluates each on-policy response against rubrics and produces a scalar score. We use the prompt template in \Cref{fig:rl_prompt}.

\begin{figure}[h]
    \begin{tcolorbox}
    You are an expert evaluator. Given a user prompt, a generated response, and a list of quality rubrics, please rate the overall quality of the response on a scale of 1 to 10 based on how well it satisfies the rubrics. \\
    Consider all rubrics holistically when determining your score. A response that violates multiple rubrics should receive a lower score, while a response that satisfies all rubrics should receive a higher score. \\ \\
    <prompt> \\
    \{instruction\} \\
    </prompt> \\ \\
    <response> \\
    \{response\} \\
    </response> \\ \\
    <rubrics> \\
    \{rubric\_list\_string\} \\
    </rubrics> \\ \\
    First, analyze the response against each rubric item, discussing how well the response meets or fails each criterion. Then, provide your final score as an integer between 1 and 10, wrapped in <score> and </score> tags. \\
    Example ending: \\
    <score> your\_integer\_score\_from\_1\_to\_10 </score> \\ \\
    Your evaluation:
    \end{tcolorbox}
    \caption{Prompt template for RL rubric-based scoring. The integer score extracted from the \texttt{<score>} tags is used as the scalar reward.}
    \label{fig:rl_prompt}
\end{figure}

\newpage
\subsection{Ablation Prompt Templates}
\label{app:ablation_templates}

Below are the prompt templates used in the ablation configurations (\Cref{tab:ablations}). Each configuration has a scoring prompt (sent to $M$) and a context template (used to condition the teacher).

\paragraph{Full Critique.} As shown in \Cref{fig:ablation_full_critique}. The scoring prompt is identical to RL (\Cref{fig:rl_prompt}). The context template prepends the entire scoring output from $M$. 

\begin{figure}[h]
    \begin{tcolorbox}
    \textbf{Scoring prompt:} Same as \Cref{fig:rl_prompt}. \\[6pt]
    \textbf{Context template:} \\ \\
    Here is an evaluation of a previous attempt on this problem: \\ \\
    {[}Evaluation by Scoring Model{]} \\
    \{scoring\_output\} \\ \\
    Now, given the above evaluation, please solve the problem with improvements: \\
    \{prompt\}
    \end{tcolorbox}
    \caption{\textit{Full Critique} ablation. The scoring prompt is the same as RL. The full textual output from $M$ (including per-rubric analysis and score) is used as context for the teacher.}
    \label{fig:ablation_full_critique}
\end{figure}

\paragraph{Rubrics Only.} As shown in \Cref{fig:ablation_rubrics_only}. The context template provides only the rubrics without any response-specific feedback.

\begin{figure}[h]
    \begin{tcolorbox}
    \textbf{Scoring prompt:} None \\[6pt]
    \textbf{Context template:} \\ \\
    Here are quality rubrics that your response will be evaluated against: \\ \\
    <rubrics> \\
    \{rubric\_list\_string\} \\
    </rubrics> \\ \\
    Please solve the following problem, keeping the above evaluation criteria in mind: \\
    \{prompt\}
    \end{tcolorbox}
    \caption{\textit{Rubrics Only} ablation. There is no scoring prompt. Only the rubrics are provided as context, with no response-specific feedback from $M$.}
    \label{fig:ablation_rubrics_only}
\end{figure}

\paragraph{Multiple-Choice.} As shown in \Cref{fig:ablation_multiple_choice}. The scoring prompt extends the RL prompt by appending an instruction to select one of 9 predefined directives. If $M$ outputs none of them, the experiential knowledge is left empty. So there are 10 context choices in total. The context template is the same as default \ours{} (\Cref{fig:el_context_prompt}), with the selected directive as the experiential knowledge.

\begin{figure}[h]
    \begin{tcolorbox}
    \textbf{Scoring prompt} (appended after the standard scoring instructions in \Cref{fig:rl_prompt}): \\ \\
    After the score, select exactly ONE of the following 9 improvement directives that best matches the primary area where the response needs improvement. Output your selection wrapped in <experience> and </experience> tags. You must output the directive exactly as written below, with no modifications. \\ \\
    1. ``Improve factual accuracy and correctness of claims.'' \\
    2. ``Follow the user's instructions and constraints more precisely.'' \\
    3. ``Provide more depth, detail, and thorough coverage of the topic.'' \\
    4. ``Improve clarity, organization, and logical flow of the response.'' \\
    5. ``Adopt a more appropriate tone, style, or register for the context.'' \\
    6. ``Be more concise and eliminate unnecessary verbosity or repetition.'' \\
    7. ``Enhance creativity, originality, or engagement in the response.'' \\
    8. ``Improve safety, sensitivity, and avoidance of harmful content.'' \\
    9. ``No significant improvement needed. Maintain current quality.'' \\ \\
    Example ending: \\
    <score> your\_integer\_score\_from\_1\_to\_10 </score> \\ \\
    <experience> \\
    SELECTED\_IMPROVEMENT \\
    </experience> \\
    Your evaluation: \\[6pt]
    \textbf{Context template:} Same as \Cref{fig:el_context_prompt}, with the selected directive as \texttt{\{experience\}}.
    \end{tcolorbox}
    \caption{\textit{Multiple-Choice} ablation. The scoring prompt appends a directive selection task. The selected directive is extracted from the \texttt{<experience>} tags and used as context with the same template as default \ours{}. If $M$ outputs none of required directives, the experiential knowledge is left empty. So there are 10 context choices in total.}
    \label{fig:ablation_multiple_choice}
\end{figure}

\subsection{Rubric Examples}
\label{app:rubric_examples}

Each training prompt is paired with a set of evaluation rubrics generated by GPT-4o. Each rubric has a title, description, and weight indicating its importance. Below are three examples spanning different task types as in \Cref{fig:rubric_examples}.

\begin{figure}[h]
    \begin{tcolorbox}
    \textbf{Prompt:} Given the current state of the world, what are the most pressing issues that humanity faces and what are some possible solutions or actions that could be taken to address them? Explain your reasoning and provide evidence or examples to support your claims. \\[6pt]
    \textbf{Rubrics:} \\
    \textbullet~\textbf{Comprehensive Issue Coverage} (weight 5): Identifies and discusses multiple pressing global issues. \\
    \textbullet~\textbf{Solution Proposals} (weight 4): Offers feasible solutions or actions for the issues discussed. \\
    \textbullet~\textbf{Evidence and Examples} (weight 4): Provides evidence or examples to support claims and solutions. \\
    \textbullet~\textbf{Reasoning Clarity} (weight 4): Explains reasoning clearly and logically for the issues and solutions presented. \\
    \textbullet~\textbf{Factual Accuracy} (weight 5): Ensures all factual statements are accurate and verifiable. \\
    \textbullet~\textbf{Empathetic Tone} (weight 2): Maintains an empathetic and understanding tone throughout the response.
    \end{tcolorbox}
    \vspace{0.3cm}
    \begin{tcolorbox}
    \textbf{Prompt:} From the following tables, write a SQL query to find the director of a movie that cast a role as Sean Maguire. Return director first name, last name and movie title. \\[6pt]
    \textbf{Rubrics:} \\
    \textbullet~\textbf{SQL Syntax Correctness} (weight 5): The response must provide a SQL query without syntax errors. \\
    \textbullet~\textbf{Correct Table and Column Usage} (weight 5): The response must demonstrate the correct usage of table and column names as specified or implied in the prompt. \\
    \textbullet~\textbf{Query Completeness} (weight 4): The SQL query must include all necessary joins and conditions to accurately find the director of the movie with the specified role. \\
    \textbullet~\textbf{Result Clarity} (weight 4): The SQL query must clearly return the director's first name, last name, and the movie title, as requested. \\
    \textbullet~\textbf{Query Efficiency} (weight 1): The SQL query should aim to be efficient, using appropriate indexing or limiting strategies where applicable.
    \end{tcolorbox}
    \vspace{0.3cm}
    \begin{tcolorbox}
    \textbf{Prompt:} We're doing a catering for 150 people. The serving portion is 2 tablespoons. I am going to give you a recipe which yields 12 portions. You will then give me the proper ratios if we're doing it for 150 people. \\[6pt]
    \textbf{Rubrics:} \\
    \textbullet~\textbf{Mathematical Accuracy} (weight 5): Calculates correct ingredient ratios for 150 servings based on the initial 12-portion recipe. \\
    \textbullet~\textbf{Clarity of Instructions} (weight 4): Provides clear and unambiguous steps for adjusting the recipe quantities. \\
    \textbullet~\textbf{Recipe Completeness} (weight 3): Includes all ingredients from the original recipe in the adjusted version. \\
    \textbullet~\textbf{Formatting} (weight 2): Presents the adjusted recipe in a well-organized and easy-to-read format. \\
    \textbullet~\textbf{Pitfall: Incorrect Scaling} (weight $-$2): Does not mention or incorrectly scales the recipe to accommodate 150 servings.
    \end{tcolorbox}
    \caption{Three rubric examples from the training set, covering an open-ended analytical question (top), a technical coding task (middle), and a recipe scaling task with a negative-weight pitfall criterion (bottom).}
    \label{fig:rubric_examples}
\end{figure}

\subsection{Experiential Knowledge Examples}
\label{app:experience_examples}

Below are three examples of experiential knowledge extracted by the LLM-as-a-Coach during training. Each example corresponds to a different type of task, as shown in \Cref{fig:experience_examples}.

\begin{figure}[h]
    \begin{tcolorbox}
    Key strategies include: \\
    1. \textbf{Maintaining Emotional Arc}: The response respects the emotional progression of the character, allowing vulnerability to slowly transition into curiosity and cautious hope. This creates a believable and engaging character experience. \\
    2. \textbf{Respecting the Prompt Context}: The response stays closely aligned with the given dialogue and scenario, focusing on the interaction without introducing unrelated or off-track information. \\
    3. \textbf{Clear and Accessible Language}: The language is simple and conversational, avoiding jargon or overly complex descriptions. This makes the world feel both immersive and understandable. \\
    4. \textbf{Safe and Supportive Tone}: The response consistently uses a comforting and supportive tone, addressing distress and offering reassurance, which aligns with the emotional needs of the character. \\
    5. \textbf{Balanced World-Building}: The response provides enough detail about the setting without overwhelming the character. It introduces the environment in a way that feels organic and grounded in the scene. \\
    6. \textbf{Avoiding Pitfalls}: The response avoids both overwhelming information and misleading directions, ensuring that the narrative remains safe and coherent. \\[4pt]
    These strategies can be applied to future responses to maintain emotional authenticity, clarity, and narrative coherence while staying true to the character and setting.
    \end{tcolorbox}
    \vspace{0.3cm}
    \begin{tcolorbox}
    When responding to factual questions, especially those involving specific measurements, it is crucial to prioritize \textbf{accuracy} and \textbf{clarity}. Always verify the exact figures and ensure that the response directly addresses the user's query without unnecessary elaboration. \\[4pt]
    To improve future responses, focus on \textbf{precision} in factual data and \textbf{directness} in addressing the question. Avoid assumptions or explanations that may introduce confusion. When providing additional context, ensure it is relevant and enhances understanding without diverting from the core question. Also, maintain a \textbf{clear structure} that separates factual data from explanatory content to improve readability and user satisfaction.
    \end{tcolorbox}
    \vspace{0.3cm}
    \begin{tcolorbox}
    To create a high-quality technical report, it is essential to thoroughly understand the subject matter and structure the content in a logical, well-organized manner. The report should be comprehensive, covering all key aspects of the topic while maintaining a clear focus and depth of analysis. Ensuring that the response adheres to the specified word count is crucial for meeting user expectations and demonstrating thoroughness. Including relevant technical details and using appropriate terminology shows a deep understanding of the subject. Maintaining correct grammar and formatting enhances professionalism and readability. Finally, ensuring that the content is accurate, safe, and compliant with technical standards is essential to avoid misinformation and ensure the report is suitable for its intended audience.
    \end{tcolorbox}
    \caption{Three examples of experiential knowledge extracted by the LLM-as-a-Coach. Each captures transferable strategies rather than instance-specific corrections.}
    \label{fig:experience_examples}
\end{figure}

\subsection{End-to-End Example}
\label{app:e2e_example}

We present a complete end-to-end example of the \ours{} pipeline in \Cref{fig:e2e_example}, showing the prompt, rubrics, on-policy response from the policy model, and the experiential knowledge extracted by the LLM-as-a-Coach.

\begin{figure}[h]
    \begin{tcolorbox}[title=Prompt]
    What is a disclaimer to put on a mental wellness app that lets others know that the relationship between the app and client is not clinical in nature and not to be a substitute for professional clinical service?
    \end{tcolorbox}
    \vspace{0.2cm}
    \begin{tcolorbox}[title=Rubrics]
    \textbullet~\textbf{Explicit Disclaimer} (weight 5): Clearly states that the app is not a substitute for professional clinical services. \\
    \textbullet~\textbf{Non-Clinical Nature} (weight 4): Specifies that the app-client relationship is not clinical in nature. \\
    \textbullet~\textbf{Professional Advice Caution} (weight 5): Advises users to seek professional clinical advice for mental health issues. \\
    \textbullet~\textbf{Legal Liability Limit} (weight 4): Limits legal liability by clarifying the app's role and scope. \\
    \textbullet~\textbf{Clear Language} (weight 3): Uses clear and understandable language to convey the disclaimer. \\
    \textbullet~\textbf{Pitfall: Non-Specific Advice} (weight $-$2): Recommends specific treatments or therapies without professional input.
    \end{tcolorbox}
    \vspace{0.2cm}
    \begin{tcolorbox}[title=On-Policy Response (truncated)]
    Here's a professional and clear disclaimer you can include on your mental wellness app: \\[4pt]
    \textbf{Disclaimer} \\
    The information and services provided by [App Name] are for general informational and educational purposes only and are not intended to replace professional mental health care. The app does not establish a clinical or therapeutic relationship with users. [App Name] is not a substitute for the advice, diagnosis, or treatment of a qualified mental health professional. \\[4pt]
    Users are encouraged to consult with a licensed mental health provider for any mental health concerns. [App Name] does not provide medical, psychological, or clinical services \ldots \\[4pt]
    \textit{[195 words total]}
    \end{tcolorbox}
    \vspace{0.2cm}
    \begin{tcolorbox}[title=Extracted Experiential Knowledge]
    1. \textbf{State limitations before capabilities}: When drafting disclaimers or boundary-setting text, lead with explicit negations (e.g., ``does not establish,'' ``is not a substitute'') to prevent misinterpretation of the service's scope. \\
    2. \textbf{Pair each limitation with a constructive redirect}: Every statement of what the service does \textit{not} provide should be followed by guidance on where to find that service (e.g., ``consult with a licensed provider''), ensuring the user is not left without a path forward. \\
    3. \textbf{Use standard legal phrasing for credibility}: Recognized formulations such as ``for general informational purposes only'' and ``does not constitute a professional relationship'' convey authority and completeness without requiring custom legal review. \\
    4. \textbf{Keep tone neutral and accessible}: Disclaimers should be written in plain language that any user can understand, avoiding both overly casual phrasing and impenetrable legalese. \\
    5. \textbf{Offer adaptation at the end}: When the user's exact deployment context is unclear, close with an offer to tailor the output (e.g., different platforms or tones), demonstrating helpfulness beyond the literal request.
    \end{tcolorbox}
    \caption{End-to-end example of the \ours{} pipeline. Given a prompt and rubrics, the policy generates a response; the LLM-as-a-Coach then extracts transferable experiential knowledge that is prepended as context to the teacher model for on-policy context distillation.}
    \label{fig:e2e_example}
\end{figure}

\end{document}